%% file: main.tex
\documentclass[11pt]{article}

\ifdefined\pdfminorversion
\fi

\usepackage[preprint]{acl}
\usepackage{times}
\usepackage{latexsym}
\usepackage[T1]{fontenc}
\usepackage[utf8]{inputenc}
\usepackage{microtype}
\usepackage{amsmath}
\usepackage{amssymb}
\usepackage{graphicx}
\usepackage{booktabs}
\usepackage{tabularx}
\usepackage{array}
\usepackage{xcolor}
\usepackage{colortbl}

\newcommand{\goldset}{\mathcal{A}^{*}}
\newcommand{\opcredit}{\emph{OpCredit}}
\newcolumntype{Y}{>{\centering\arraybackslash}X}

\title{OpCredit: Operation-Aware Policy Learning for Agentic Knowledge Graph Question Answering}

\author{
  Yucheng Li$^{1}$ \quad Mingyuan Li$^{1}$ \quad Xiaxia Wang$^{2}$ \quad
  Gong Cheng$^{1*}$ \\
  $^{1}$State Key Laboratory of Novel Software Technology, Nanjing University, Nanjing, China \\
  $^{2}$Department of Computer Science, University of Oxford \\
  $^{*}$Corresponding author: \texttt{gcheng@nju.edu.cn}
}

\hypersetup{
  pdftitle={OpCredit: Operation-Aware Policy Learning for Agentic Knowledge Graph Question Answering},
  pdfauthor={Yucheng Li, Mingyuan Li, Xiaxia Wang, Gong Cheng}
}

\begin{document}
\maketitle

\begin{abstract}
Agentic knowledge graph question answering (KGQA) relies on sequences of executable operations, but trajectory-level rewards obscure their individual contributions.
Successful trajectories may contain ineffective operations, while failed trajectories may contain useful ones.
To assess individual operations, we propose \emph{OpCredit}, an operation-aware policy-learning framework that combines downstream outcome credit with immediate transition credit:
Branch Outcome Credit (BOC) reuses ordinary rollouts to compare terminal task rewards observed after different operations under matched navigation contexts;
Frontier Transition Credit (FTC) measures operation-induced changes in answer-conditioned frontier potential.
The two complementary signals jointly supervise the corresponding decision spans during training, while inference uses only the learned policy and standard KG interface.
Experiments on ComplexWeb\-Questions, WebQSP, and GrailQA across three backbones demonstrate consistent improvements in KGQA performance.
\end{abstract}

\begin{figure}[!t]
\centering
\includegraphics[width=0.95\columnwidth]{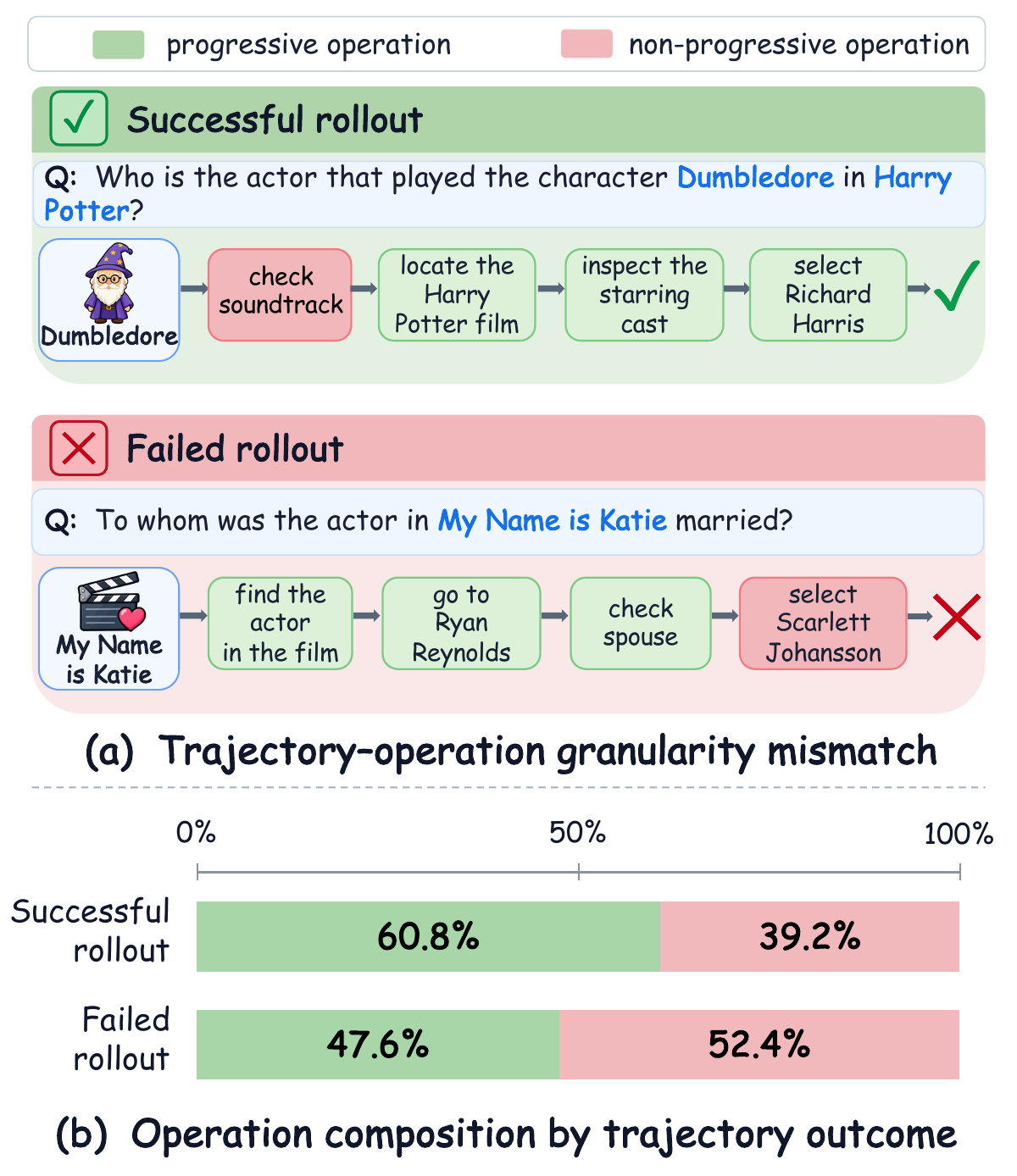}
\caption{Trajectory--operation granularity mismatch in KGQA. Progressive operations advance toward the gold answer. (a) Examples in successful and failed rollouts. (b) Operation proportions on CWQ with SFT-initialized Qwen2.5-7B.}
\label{fig:intro_mismatch}
\end{figure}

\section{Introduction}

Knowledge graph question answering (KGQA) grounds answers in structured facts and offers an explicit reasoning substrate for knowledge-intensive questions.
Recent agentic approaches answer complex questions through multi-turn interaction between a large language model (LLM) and a knowledge graph (KG): the policy issues executable operations, observes the returned evidence, and decides how to continue navigating or when to answer~\citep{tog,pog,kgr1}.
Each operation determines which evidence becomes available to subsequent decisions.
The choice of local operations therefore shapes the evidence on which successful reasoning depends.

Final-answer rewards provide a natural way to improve such agents without annotating every intermediate decision, but they evaluate an entire trajectory at once: a successful rollout may contain off-target operations, while a failed rollout may contain useful navigation.
Group Relative Policy Optimization (GRPO) compares same-question rollouts and applies each trajectory's advantage to all policy-generated tokens~\citep{grpo}, so it does not explicitly distinguish operations that advance navigation from those that stall or divert it.
Figure~\ref{fig:intro_mismatch}(a) illustrates this mismatch: the successful rollout includes an off-target soundtrack lookup before recovering to the correct actor, whereas the failed rollout reaches a relevant actor but ends with an incorrect answer; the diagnostic in Figure~\ref{fig:intro_mismatch}(b) likewise finds many progressive operations in failed rollouts and non-progressive operations in successful rollouts.
These cases expose a \emph{trajectory--operation granularity mismatch}: the outcome is trajectory-level, while the policy must select operations individually.

Recent agentic reinforcement learning (RL) methods redistribute trajectory feedback across local decisions by grouping repeated states or aggregating sampled rollout structures~\citep{gigpo,rtmc,graphgpo}.
Prior KGQA work provides intermediate guidance through reference-path supervision, learned process evaluators, or graph-based progress signals~\citep{eog,dprm,dtpo}.
Specifically for agentic KGQA, \emph{a natural approach is to assess local operations through their observed downstream outcomes}.

This motivates our \emph{Branch Outcome Credit (BOC)}.
BOC organizes same-question rollouts post hoc by navigation context and compares terminal task rewards observed after different operations.
These context-relative comparisons assign outcome credit to individual operations under the sampled policy, without constructing an additional rollout tree or training an auxiliary evaluator.

BOC captures the downstream consequence of an operation, but later decisions can obscure the operation's immediate effect on the KG evidence frontier.
To address it, we introduce \emph{Frontier Transition Credit (FTC)}.
FTC evaluates the immediate effect of each operation by comparing the KG evidence frontier's proximity to the answer before and after execution.

Bringing these two views together, we propose \emph{OpCredit}, an operation-aware policy-learning framework for agentic KGQA.
BOC and FTC jointly supervise the corresponding decision spans to update a single KGQA policy.
At inference time, the policy uses the standard KG interface without training-time signals or an auxiliary evaluator.

Experiments on ComplexWebQuestions (CWQ), WebQSP, and GrailQA with three LLM backbones show consistent gains in KGQA performance.
Controlled comparisons show that BOC and FTC each improve on GRPO, with their combination performing best.
A reference-blinded operation-level audit finds closer alignment between assigned credit and assessed usefulness for OpCredit than for GRPO.

Our contributions are twofold:
\begin{itemize}
    \item We propose OpCredit, an operation-aware policy-learning framework for agentic KGQA that addresses the trajectory--operation granularity mismatch through complementary outcome and transition credit.
    \item We instantiate this framework with BOC and FTC. BOC compares terminal task rewards observed after different operations under matched navigation contexts, while FTC measures signed changes in answer-conditioned frontier potential.
\end{itemize}

Our code is available at \url{https://github.com/kaka1215/OpCredit}.

\section{Related Work}

\subsection{LLM-Based KGQA}

For training-free KG exploration, ToG uses LLM-guided beam search~\citep{tog}, while PoG adds adaptive planning and self-correction~\citep{pog}.
Learning-based methods learn path planning, evidence use, or graph interaction~\citep{rog,kgr1,eog}.
\emph{We focus on operation-aware policy learning for agents that select executable operations during multi-turn KG interaction.}

\subsection{Supervision for KGQA}

Task outcomes and interaction feedback provide supervision from answers and graph access.
KnowCoder-A1 trains agentic reasoning with a curriculum of trajectory-level outcome rewards~\citep{knowcodera1}.
KG-R1 combines global answer and retrieval rewards with turn-wise validity feedback~\citep{kgr1}.
\emph{However, terminal task rewards require further attribution to distinguish individual operations, while interaction validity alone does not establish an operation's contribution to reasoning.}

Reference trajectories and learned evaluators provide more explicit guidance for reasoning.
GraphWalker learns KG exploration and error recovery through two-stage supervised fine-tuning on synthetic and expert trajectories before outcome-based RL~\citep{graphwalker}.
EoG rewards agreement with verified reference paths~\citep{eog}. GAPD aligns rollout and gold states for gold-action-conditioned token-level self-distillation~\citep{gapd}.
DPRM learns dual implicit process reward models (PRMs) from outcome supervision to score KG and chain-of-thought (CoT) steps for inference-time selection~\citep{dprm}.
\emph{Such supervision remains constrained by the coverage of demonstrations, reference paths, gold actions, or by the generalization capabilities of learned evaluators.}

Graph-based progress uses the task KG to assess intermediate decisions.
DTPO uses answer-distance changes and downstream outcomes to supervise KG operations sampled through tree expansion~\citep{dtpo}.
\emph{Its local comparisons require explicit branch expansion and KG execution, making rollout cost sensitive to branching width.}

\subsection{Credit Assignment in Agentic RL}

Agentic RL methods also derive local credit from shared structure in sampled rollouts. Table~\ref{tab:supervision_capabilities} compares outcome attribution and process signals across representative methods.
GiGPO augments episode-level advantages through action grouping at repeated environment states~\citep{gigpo}.
RTMC builds rollout trees from compact state--action signatures to estimate step-level advantages~\citep{rtmc}.
GraphGPO combines a trajectory-level outcome advantage with progress measured by distances to successful states in a sampled experience graph~\citep{graphgpo}.
These approaches derive local credit from relationships among sampled states and outcomes. \emph{Our OpCredit uses such outcome evidence for BOC and further evaluates immediate transitions in the task KG through FTC. The distinction lies in aligning rollout-based outcome credit and KG-grounded transition credit with the same operation's decision span.}

\begin{table}[t]
\centering
\footnotesize
\setlength{\tabcolsep}{3pt}
\renewcommand{\arraystretch}{1.12}
\begin{tabularx}{\columnwidth}{
@{}
> {\raggedright\arraybackslash}p{2.2cm}
> {\raggedright\arraybackslash}p{2.2cm}
> {\raggedright\arraybackslash}X
@{}}
\toprule
\textbf{Method}
& \textbf{Outcome attribution}
& \textbf{Process signal} \\
\midrule
\rowcolor{gray!12}
\multicolumn{3}{l}{\textit{Supervision for KGQA}} \\
KnowCoder-A1 & Trajectory & -- \\
KG-R1 & Trajectory & Turn validity \\
GraphWalker & Trajectory & Trajectory SFT \\
EoG & Trajectory & Reference-path matching \\
GAPD & Trajectory & Gold-action distillation \\
DPRM & Local\newline (implicit) & KG/CoT step scores \\
DTPO & Local & Answer distance \\
\midrule
\rowcolor{gray!12}
\multicolumn{3}{l}{\textit{Credit Assignment in Agentic RL}} \\
GiGPO & Local & Validity penalties \\
RTMC & Local & Validity / validation rewards \\
GraphGPO & Trajectory & Experience-graph distance \\
\midrule
\rowcolor{blue!10}
\textbf{OpCredit}
& Local\newline (context-relative)
& Frontier transition\newline (answer-conditioned) \\
\bottomrule
\end{tabularx}
\par\smallskip
\parbox{\columnwidth}{\footnotesize
Trajectory/Local: attribution to complete trajectories/intermediate decisions; --: no corresponding signal. Validity and validation rewards are task-dependent.}
\caption{Comparison of outcome attribution and process signals across representative methods.}
\label{tab:supervision_capabilities}
\end{table}

\section{Method}
\label{sec:method}

\begin{figure*}[t]
\centering
\includegraphics[width=\textwidth]{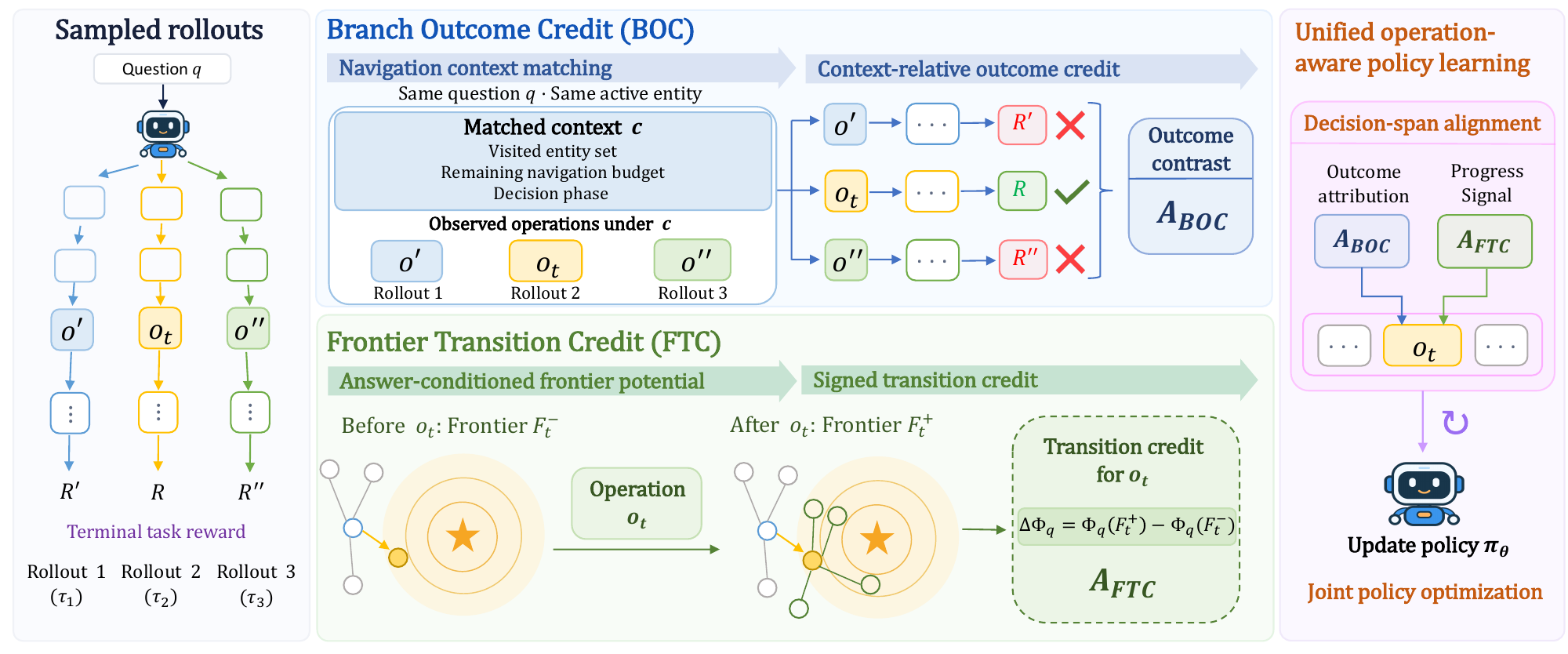}
\caption{Overview of OpCredit. BOC compares terminal task rewards under matched navigation contexts; FTC measures answer-conditioned changes in frontier potential. Both signals supervise the same decision span during training. Inference uses the learned policy and standard KG interface.}
\label{fig:overview}
\end{figure*}

OpCredit supervises individual KG operations through two complementary signals: Branch Outcome Credit (BOC) compares terminal task rewards under matched navigation contexts, while Frontier Transition Credit (FTC) measures immediate changes in answer-conditioned frontier potential.
Both signals supervise the same decision span to update a single policy (Figure~\ref{fig:overview}).

\subsection{Preliminaries}
\label{sec:preliminaries}

Let $\mathcal{K}=(\mathcal{V},\mathcal{E})$ be a knowledge graph, $q$ a question, and $\goldset(q)\subseteq\mathcal{V}$ the set of gold answer entities mapped to the KG for training.
An agent interacts with $\mathcal{K}$ through operations such as \emph{entity-and-direction selection}, \emph{relation selection}, and \emph{answer emission}.
At step $t$, the policy observes a navigation state $s_t$ and selects an operation $o_t\in\mathcal{O}(s_t)$, where $\mathcal{O}(s_t)$ denotes the available operations.
Navigation operations return graph observations that update the state to $s_{t+1}$, while answer emission terminates the interaction.
A trajectory $\tau=(s_0,o_0,\ldots,s_{T-1},o_{T-1},s_T)$ consists of $T$ operations, including the terminal answer decision.

We initialize the policy through supervised fine-tuning (SFT) to generate reasoning followed by executable operations, conditioned on the question and interaction history.
Appendix~\ref{app:templates} details the interaction protocol, template-based SFT trajectory construction, and serialized examples.
At each training iteration, the rollout policy $\pi_{\theta_{\mathrm{old}}}$ samples $N$ trajectories $\{\tau_i\}_{i=1}^{N}$ for the same question through the standard KG interface.
Each trajectory $\tau_i$ receives a terminal task reward $R_{\mathrm{task}}(\tau_i)$ based on its final-answer outcome; Appendix~\ref{app:hparams} specifies the reward and training hyperparameters.
GRPO computes a trajectory-level advantage within this group:
\begin{equation}
A_i^{\mathrm{traj}}
=\frac{R_{\mathrm{task}}(\tau_i)-\mu_q}{\sigma_q+\varepsilon},
\label{eq:traj_advantage}
\end{equation}
where $\mu_q$ and $\sigma_q$ are the mean and standard deviation of terminal task rewards within the group, and $\varepsilon>0$ is a small constant for numerical stability.
GRPO assigns this advantage to all policy-generated tokens in $\tau_i$.

\subsection{Branch Outcome Credit (BOC)}
\label{sec:branch_credit}

BOC assigns context-relative outcome credit to individual operations by comparing terminal task rewards under matched navigation contexts.

\paragraph{Navigation context matching.}
For same-question rollouts with $R_i=R_{\mathrm{task}}(\tau_i)$, BOC first groups decisions by the active entity before each operation.
Within each group, the pre-decision navigation state $s$ is mapped to
\begin{equation}
\label{eq:context_key}
c=\chi(s)=\bigl(E_{\mathrm{vis}}(s), b(s), \phi(s)\bigr),
\end{equation}
where $E_{\mathrm{vis}}(s)$ is the visited entity set, $b(s)$ the remaining navigation budget, and $\phi(s)$ the decision phase: relation selection, entity-and-direction selection, or answer emission.
Here $b(s)=H-h(s)$, with maximum hop budget $H$ and $h(s)$ completed hops; decisions within a hop share the remaining budget.
Matching these context elements helps to pool outcome evidence and keep rollout comparisons within the same decision phase, without additional environment interaction.
Let $\Omega_q(c,o)$ contain the unique rollout indices observing operation $o$ under context $c$, and $\Omega_q(c)=\bigcup_o\Omega_q(c,o)$.
Both support sets are restricted to the same active entity, omitted from the notation.
Each rollout contributes once per set, avoiding duplicate outcome evidence.

\paragraph{Context-relative outcome credit.}
From sampled continuations, we estimate the operation's observed return and context baseline by
\begin{equation}
\label{eq:branch_qv}
\begin{aligned}
\widehat Q_{\mathrm{out}}(c,o)
&=\frac{1}{|\Omega_q(c,o)|}\sum_{i\in\Omega_q(c,o)}R_i,\\
\widehat V_{\mathrm{out}}(c)
&=\frac{1}{|\Omega_q(c)|}\sum_{i\in\Omega_q(c)}R_i.
\end{aligned}
\end{equation}
Here $\widehat Q_{\mathrm{out}}$ averages returns after operation $o$, while $\widehat V_{\mathrm{out}}$ averages returns over rollouts encountering a matched context; both include successful and unsuccessful completions.
Unlike the question-level mean $\mu_q$ in Eq.~\eqref{eq:traj_advantage}, $\widehat V_{\mathrm{out}}(c)$ is context-specific.
For decision $t$ in rollout $\tau_i$, let $\Delta_{i,t}=\widehat Q_{\mathrm{out}}(c_{i,t},o_{i,t})-\widehat V_{\mathrm{out}}(c_{i,t})$ denote the observed outcome contrast under context $c_{i,t}$.
BOC uses the normalized contrast when nonzero and otherwise retains the trajectory-level advantage in Eq.~\eqref{eq:traj_advantage}:
\begin{equation}
\label{eq:branch_advantage}
A_{\mathrm{BOC}}^{i,t}
=\begin{cases}
\dfrac{\Delta_{i,t}}{\sigma_q+\varepsilon}, & \Delta_{i,t}\ne 0,\\[3pt]
A_i^{\mathrm{traj}}, & \Delta_{i,t}=0
\end{cases}.
\end{equation}
The contrast's sign indicates whether the operation's observed return is above or below its context baseline; both cases use the same question-level reward scale.
In Figure~\ref{fig:overview}, $R'=R''=0$ and $R=1$ yield positive BOC for $o_t$ and negative BOC for $o',o''$; Appendix~\ref{app:boc_example} gives the calculation.

\subsection{Frontier Transition Credit (FTC)}
\label{sec:frontier_transition}

FTC evaluates the effect of each operation on the KG evidence frontier through an answer-conditioned before/after comparison, without requiring a reference navigation sequence.

\paragraph{Answer-conditioned frontier potential.}
For operation $o_{i,t}$ at decision $t$ in trajectory $\tau_i$, let $F_{i,t}^{-}$ and $F_{i,t}^{+}$ denote the operation-local entity sets before and after its transition, respectively.
Figure~\ref{fig:overview} omits the rollout index $i$.
A useful expansion may expose an answer-relevant candidate alongside distractors, so we assess the potential of a nonempty frontier $F$ by its best available opportunity:
\begin{equation}
\label{eq:frontier_potential}
\Phi_q(F)=\max_{e\in F}\bigl(2-\delta_q(e)\bigr).
\end{equation}
Here $\delta_q(e)=0$ for a gold answer entity, $1$ for an entity outside the gold answer set but in its constructed one-hop neighborhood, and $2$ for all remaining entities.
These labels describe coarse answer proximity, not exact graph distance or remaining operation counts.
Thus $\Phi_q$ measures frontier-to-answer proximity.
The gold answer set is used only during training and is not added to policy observations; frontier definitions by operation, neighborhood construction, and unavailable-frontier handling appear in Appendix~\ref{app:ftc_details}.

\paragraph{Signed transition credit.}
The potential of a resulting frontier alone does not distinguish progress made by the current operation from proximity already achieved by earlier decisions.
We therefore assign the operation the signed change between its before/after frontier potentials:
\begin{equation}
\label{eq:structural_advantage}
A_{\mathrm{FTC}}^{i,t}
=\Phi_q(F_{i,t}^{+})-\Phi_q(F_{i,t}^{-}).
\end{equation}
Positive, zero, and negative values indicate measured progress, no measured change, and regression, respectively.
The comparison uses operation-local frontiers, not the cumulative interaction history, so earlier answer encounters do not automatically make later operations favorable.
The potential-difference form is related to potential-based reward shaping~\citep{ng1999policy}; here it supplies an answer-conditioned auxiliary objective on decision spans rather than changing the terminal task reward used for BOC.
An operation can therefore receive positive FTC even if later decisions fail, while BOC evaluates its sampled continuations through terminal task rewards.

\subsection{Unified Operation-Aware Policy Learning}
\label{sec:policy_learning}

BOC and FTC are constructed from rollouts and aligned to the same decision span for a joint policy update.

\paragraph{Decision-span alignment.}
Let $y_i=(y_{i,1},\ldots,y_{i,L_i})$ be the policy-generated token sequence for rollout $i$, where $L_i$ is its length and $j$ indexes token positions.
Each decision comprises the generated reasoning together with its executable operation; both belong to the same decision span and receive its operation-level supervision.
For operation $o_{i,t}$, let $\mathcal J_{i,t}$ contain the token indices of this span.
The decision spans, including the terminal answer decision, partition the generated sequence: $\mathcal I_i=\{1,\ldots,L_i\}=\bigcup_t\mathcal J_{i,t}$.
For $j\in\mathcal J_{i,t}$, the outcome credit is $A_{\mathrm{BOC}}^{i,t}$ from Eq.~\eqref{eq:branch_advantage}.
For $j\in\mathcal J_{i,t}$, we set $B_{i,j}=A_{\mathrm{FTC}}^{i,t}$ when the operation has a valid before/after frontier comparison; tokens without an applicable FTC signal have $B_{i,j}=0$.
The FTC eligibility rules are given in Appendix~\ref{app:ftc_details}.

\paragraph{Joint policy optimization.}
Let $h_{i,j}$ denote the full policy context preceding token $y_{i,j}$, including the question, interaction history, and previously generated tokens.
The policy ratio is
\begin{equation}
\label{eq:policy_ratio}
\rho_{i,j}(\theta)=
\frac{\pi_\theta(y_{i,j}\mid h_{i,j})}
{\pi_{\theta_{\mathrm{old}}}(y_{i,j}\mid h_{i,j})}.
\end{equation}
The BOC objective uses a clipped GRPO surrogate~\citep{grpo}, with dual clipping for negative advantages. Writing the dual-clipped surrogate defined in Appendix~\ref{app:joint_details} as $D$, we obtain
\begin{equation}
\label{eq:outcome_objective}
\mathcal L_{\mathrm{BOC}}(\theta)
=-\mathbb E_{\substack{1\le i\le N\\j\in\mathcal I_i}}
\left[D\!\left(\rho_{i,j},A_{\mathrm{BOC}}^{i,t}\right)\right].
\end{equation}
Here $t$ denotes the decision whose span contains token $j$, i.e., $j\in\mathcal J_{i,t}$.
FTC supplies signed auxiliary log-likelihood supervision on the same decision span:
\begin{equation}
\label{eq:structural_objective}
\mathcal L_{\mathrm{FTC}}(\theta)
=-\mathbb E_{\substack{1\le i\le N\\j\in\mathcal I_i}}
\left[B_{i,j}\log\pi_\theta(y_{i,j}\mid h_{i,j})\right].
\end{equation}
Both objectives average over generated tokens within each trajectory and then over trajectories.
Positive FTC reinforces the corresponding generated decision, whereas negative FTC suppresses it.
FTC supplies transition supervision alongside BOC's outcome credit, and we jointly optimize
\begin{equation}
\label{eq:joint_objective}
\mathcal L(\theta)
=\mathcal L_{\mathrm{BOC}}(\theta)
+\lambda_{\mathrm{aux}}\mathcal L_{\mathrm{FTC}}(\theta)
+\mathcal L_{\mathrm{reg}}(\theta),
\end{equation}
where $\lambda_{\mathrm{aux}}$ weights FTC and $\mathcal L_{\mathrm{reg}}$ contains KL and entropy regularization.
Credits are computed from collected rollouts and held fixed during their policy updates; subsequent iterations refresh the samples and their supervision.
At inference, only the learned policy and standard KG interface are required; rollout aggregation and answer-conditioned frontier measurements are training-time procedures.

\section{Experiments}

\begin{table*}[t]
\centering
\footnotesize
\setlength{\tabcolsep}{3.7pt}
\renewcommand{\arraystretch}{1.03}
\begin{tabular*}{\textwidth}{@{\extracolsep{\fill}}llcccccc@{}}
\toprule
\textbf{Method} & \textbf{Backbone} & \multicolumn{2}{c}{\textbf{CWQ}} & \multicolumn{2}{c}{\textbf{WebQSP}} & \multicolumn{2}{c}{\textbf{GrailQA}} \\
\cmidrule(lr){3-4}\cmidrule(lr){5-6}\cmidrule(lr){7-8}
& & \textbf{Hit@1} & \textbf{F1} & \textbf{Hit@1} & \textbf{F1} & \textbf{Hit@1} & \textbf{F1} \\
\midrule
\rowcolor{gray!12}\multicolumn{8}{l}{\textit{Training-Free KGQA}} \\
Direct prompting & Qwen2.5-3B & 22.0 & 17.7 & 44.6 & 30.3 & 13.8 & 11.6 \\
Direct prompting & Qwen2.5-7B & 25.7 & 20.7 & 50.9 & 33.2 & 15.9 & 13.6 \\
Direct prompting & Llama-3.1-8B & 28.1 & 22.4 & 55.5 & 34.8 & 21.3 & 17.7 \\
ToG~\citep{tog} & GPT-4 & 68.5 & -- & 82.6 & -- & 81.4 & -- \\
PoG~\citep{pog} & GPT-4 & 75.0 & -- & 87.3 & -- & 84.7 & -- \\
ReKnoS~\citep{reknos} & GPT-4o-mini & 66.8 & -- & 83.8 & -- & 80.5 & -- \\
\midrule
\rowcolor{gray!12}\multicolumn{8}{l}{\textit{Training-Based KGQA}} \\
GNN-RAG~\citep{gnnrag} & Llama-2-7B & 66.8 & 59.4 & 85.7 & 71.3 & -- & -- \\
KG-Agent~\citep{kgagent} & Llama-2-7B & 72.2 & 69.8 & 83.3 & 81.0 & -- & 86.1 \\
KG-R1~\citep{kgr1} & Qwen2.5-3B & 65.3 & -- & 82.8 & -- & -- & -- \\
\rowcolor{blue!4}\textit{OpCredit}$_{\mathrm{SFT}}$ & Qwen2.5-3B & 71.95 & 66.02 & 85.17 & 74.96 & 81.23 & 74.07 \\
\rowcolor{blue!10}\textit{OpCredit} & Qwen2.5-3B & \textbf{76.56} & \textbf{69.31} & \textbf{88.86} & \textbf{77.72} & \textbf{86.35} & \textbf{79.17} \\
\addlinespace[1pt]
Graph-RFT~\citep{graphrft} & Qwen2.5-7B & 78.4 & -- & 87.3 & -- & 82.7 & -- \\
EoG~\citep{eog} & Qwen2.5-7B & 82.7 & 73.8 & 90.7 & 78.1 & \textbf{91.7} & \textbf{88.5} \\
\rowcolor{blue!4}\textit{OpCredit}$_{\mathrm{SFT}}$ & Qwen2.5-7B & 72.15 & 66.06 & 86.28 & 76.38 & 83.73 & 76.77 \\
\rowcolor{blue!10}\textit{OpCredit} & Qwen2.5-7B & \textbf{83.47} & \textbf{74.58} & \textbf{92.13} & \textbf{82.97} & 90.18 & 83.42 \\
\addlinespace[1pt]
KnowCoder-A1~\citep{knowcodera1} & Qwen2.5-Coder-7B & -- & 68.3 & -- & 77.2 & -- & 80.5 \\
\addlinespace[1pt]
KBQA-o1~\citep{kbqao1} & Llama-3.1-8B & -- & -- & -- & 59.8 & -- & 78.5 \\
DPRM~\citep{dprm} & Llama-3.1-8B & 73.0 & 59.9 & 87.2 & 72.3 & -- & -- \\
SCPRM~\citep{scprm} & Llama-3.1-8B & 73.3 & -- & -- & -- & -- & -- \\
DoG~\citep{dog} & Llama-3.1-8B & 76.2 & -- & 91.4 & -- & -- & -- \\
EoG~\citep{eog} & Llama-3.1-8B & \textbf{86.6} & \textbf{77.9} & 92.8 & 81.3 & \textbf{92.1} & \textbf{90.6} \\
GAPD~\citep{gapd} & Llama-3.1-8B & -- & -- & -- & \textbf{86.1} & -- & 89.5 \\
\rowcolor{blue!4}\textit{OpCredit}$_{\mathrm{SFT}}$ & Llama-3.1-8B & 73.32 & 66.62 & 87.84 & 76.54 & 86.40 & 81.63 \\
\rowcolor{blue!10}\textit{OpCredit} & Llama-3.1-8B & 84.04 & 74.03 & \textbf{93.10} & 83.86 & 91.25 & 84.33 \\
\bottomrule
\end{tabular*}
\caption{Overall Hit@1 and F1 on CWQ, WebQSP, and GrailQA. Bold highlights the best Hit@1 and F1 among methods using each OpCredit backbone. -- denotes an unreported metric.}
\label{tab:main_results}
\end{table*}

\subsection{Experimental Setup}

\paragraph{Datasets and metrics.}
We evaluate on the CWQ~\citep{cwq} and WebQSP~\citep{webqsp} test splits and the full official GrailQA~\citep{grailqa} development split.  For our models, we report F1 over normalized answer sets and Hit@1; experimental and dataset details are in Appendices~\ref{app:protocol} and~\ref{app:data_stats}, respectively.

\paragraph{Baselines.}
Table~\ref{tab:main_results} compares two groups: \emph{training-free KGQA} (direct prompting, ToG, PoG, and ReKnoS) and \emph{training-based KGQA} (GNN-RAG, KG-Agent, KG-R1, Graph-RFT, EoG, KnowCoder-A1, KBQA-o1, DPRM, SCPRM, DoG, and GAPD). Named KGQA baseline scores come from published reports; direct prompting uses our evaluation protocol. \textit{OpCredit}$_{\mathrm{SFT}}$ denotes the SFT-initialized policy.

\paragraph{Implementation.}
We experiment with three backbones: Qwen2.5-3B, Qwen2.5-7B, and Llama-3.1-8B. All RL configurations use the corresponding SFT-initialized policy and same-question rollout groups. Appendix~\ref{app:protocol} gives the training, rollout, and evaluation settings.

\subsection{Main Results}

Table~\ref{tab:main_results} shows generally stronger results for training-based KGQA. OpCredit improves Hit@1 and F1 over SFT in all nine dataset--backbone settings, including GrailQA with Qwen2.5-3B. This smaller model also exceeds all training-free baselines on Hit@1. Gains persist on WebQSP despite its higher initial scores.

OpCredit exceeds KG-R1 on CWQ/WebQSP and Graph-RFT on all three datasets. It exceeds EoG on WebQSP and on CWQ with Qwen2.5-7B, but trails on GrailQA and on CWQ with Llama-3.1-8B. GAPD also reports higher F1 on WebQSP/GrailQA with Llama-3.1-8B. Note that the comparisons with EoG and GAPD may not be entirely fair, as they both use additional resources for process supervision. EoG uses Gemini-2.5-Flash-generated demonstrations and model-verified reference paths, while GAPD uses gold-action-conditioned self-distillation~\citep{eog,gapd}; OpCredit requires only the benchmark training data and task KG, using gold answers to construct RL supervision.

\subsection{Ablation Study}

Table~\ref{tab:ablation} compares four RL configurations on Qwen2.5-7B with the same SFT initialization and training and evaluation protocols. Removing both BOC and FTC gives the GRPO baseline. Adding BOC improves all six metrics over GRPO with the same terminal task rewards. Its gains also persist when added to FTC, indicating that context-relative outcome comparisons contribute to policy learning alongside frontier-transition supervision. FTC alone yields larger gains than BOC alone in this setting, and adding it to BOC further improves both metrics on every dataset. Immediate frontier changes therefore provide useful supervision beyond observed terminal task rewards. The full method performs best across the three datasets, supporting the complementary contributions of the two signals to operation selection.

\subsection{Analysis of Navigation Context Matching}
\label{sec:context_key}

BOC compares each operation's observed return with its comparison group's outcome baseline. Relation selection and entity-and-direction selection can share the same active entity, visited entity set, and remaining navigation budget. With rollouts and terminal task rewards fixed, we compare the full context against three alternatives: active entity only, visited entity set only, and visited entity set with remaining navigation budget. All comparisons use the same question and active entity. Without decision phase, 46--48\% of decision occurrences across checkpoints belong to groups containing both operation types, pooling outcome evidence from both relation and entity-and-direction choices. Matching the decision phase provides a type-consistent baseline: relation choices are evaluated relative to relation choices, and entity-and-direction choices relative to their own type. This aligns BOC's local outcome comparison with the kind of decision the policy is making. Appendix~\ref{app:context_key} gives the matching rules and mixing statistics.

\begin{table*}[t]
\centering
\footnotesize
\setlength{\tabcolsep}{5pt}
\renewcommand{\arraystretch}{1.08}
\begin{tabular*}{\textwidth}{@{\extracolsep{\fill}}lcc*{6}{c}@{}}
\toprule
\textbf{Configuration} & \textbf{BOC} & \textbf{FTC} & \multicolumn{2}{c}{\textbf{CWQ}} & \multicolumn{2}{c}{\textbf{WebQSP}} & \multicolumn{2}{c}{\textbf{GrailQA}} \\
\cmidrule(lr){4-5}\cmidrule(lr){6-7}\cmidrule(lr){8-9}
 & & & \textbf{Hit@1} & \textbf{F1} & \textbf{Hit@1} & \textbf{F1} & \textbf{Hit@1} & \textbf{F1} \\
\midrule
\textit{OpCredit}$_{\mathrm{SFT}}$ & -- & -- & 72.15 & 66.06 & 86.28 & 76.38 & 83.73 & 76.77 \\
\textit{OpCredit}$_{-\mathrm{BOC},-\mathrm{FTC}}$ & -- & -- & 79.43 & 71.57 & 89.44 & 78.98 & 87.17 & 79.62 \\
\textit{OpCredit}$_{-\mathrm{FTC}}$ & \checkmark & -- & 80.65 & 72.15 & 90.27 & 79.09 & 88.03 & 81.49 \\
\textit{OpCredit}$_{-\mathrm{BOC}}$ & -- & \checkmark & 82.41 & 73.70 & 91.30 & 81.48 & 89.58 & 82.36 \\
\rowcolor{blue!10}\textit{OpCredit} & \checkmark & \checkmark & \textbf{83.47} & \textbf{74.58} & \textbf{92.13} & \textbf{82.97} & \textbf{90.18} & \textbf{83.42} \\
\bottomrule
\end{tabular*}
\caption{Ablation of OpCredit with Qwen2.5-7B.}
\label{tab:ablation}
\end{table*}

\subsection{Reference-Blinded Credit Audit}
\label{sec:credit_audit}

We evaluate whether operation-level credit aligns with operation usefulness through a reference-blinded audit. We implement the credit-assignment rules of GiGPO, RTMC, and GraphGPO for KGQA and replay their signals alongside GRPO and OpCredit on a common rollout pool generated by the SFT policy. With operations and terminal task rewards fixed, the audit compares credit assignments to the same decisions. Given the question, interaction history preceding the operation, the operation itself, and the graph observations it returns, an LLM judge assigns a usefulness rating without access to gold answers, terminal task rewards, or assigned credit. We report mean credit across four ordered usefulness levels and the area under the receiver operating characteristic curve (AUC), using assigned credit to distinguish operations judged useful from those judged unhelpful. Appendix~\ref{app:judge} provides the complete protocol and results.

\begin{figure}[!t]
\centering
\includegraphics[width=\columnwidth]{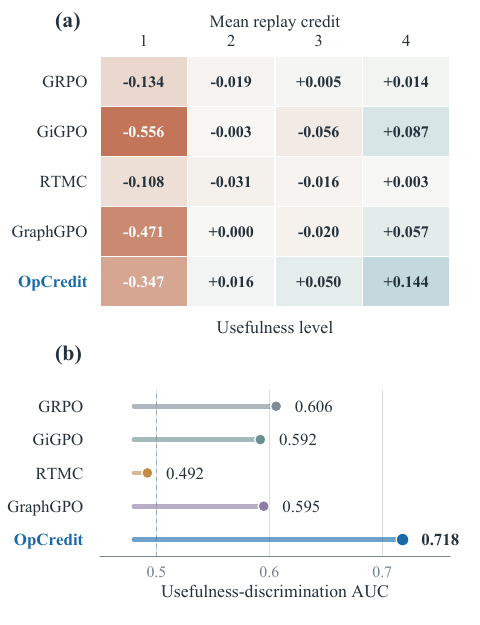}
\caption{Credit alignment on a common SFT rollout pool. (a) Mean replay credit across four ordered usefulness levels (1 is least useful and 4 is most useful). (b) Usefulness-discrimination AUC; the dashed line marks random discrimination.}
\label{fig:credit_alignment}
\end{figure}

Figure~\ref{fig:credit_alignment}(a) shows that OpCredit assigns negative mean credit to the least useful operations and increasingly positive credit across levels 2--4. GRPO and RTMC also exhibit increasing mean credit, whereas GiGPO and GraphGPO reverse this ordering between the middle levels. To compare usefulness discrimination beyond these group averages, Figure~\ref{fig:credit_alignment}(b) reports AUC based on operation-level credit rankings. OpCredit achieves the highest AUC among the compared methods (0.718), ahead of GRPO (0.606) and RTMC (0.492). These results indicate OpCredit's stronger discriminability between useful and unhelpful operations within the same rollout pool.

\begin{figure}[t]
\centering
\includegraphics[width=\columnwidth]{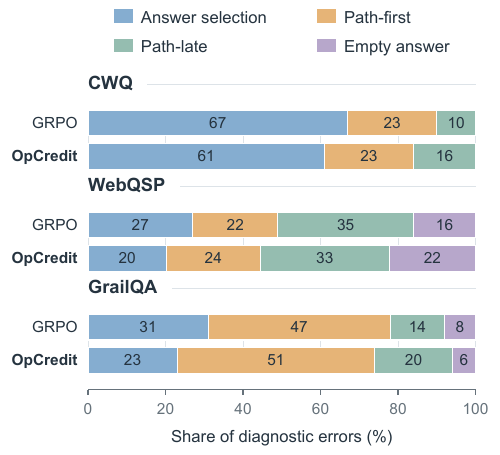}
\caption{Error composition on matched diagnostic subsets. Labels show rounded percentages within each method's error set.}
\label{fig:error_analysis}
\end{figure}

\subsection{Error Analysis}
\label{sec:error_analysis}

Figure~\ref{fig:error_analysis} compares the composition of GRPO and OpCredit errors on matched diagnostic subsets across three datasets. We distinguish answer-selection errors despite answer-relevant evidence, first-hop path deviations (Path-first), later deviations or failures without reference-path alignment (Path-late), and empty predicted answer sets (Empty answer). Answer-selection errors dominate CWQ, whereas first-hop deviations dominate GrailQA. WebQSP exhibits a more mixed error profile, with substantial shares of Path-late and Empty answer cases. The same terminal failure label can thus reflect difficulties in acquiring useful evidence or selecting answers from available evidence.

Compared with GRPO, OpCredit has a smaller share of answer-selection errors on all three datasets. Path-late accounts for a larger share of its remaining errors on CWQ and GrailQA, while Empty answer accounts for a smaller share on GrailQA. These profiles highlight navigation as a prominent source of the remaining errors. They complement the credit audit by locating failures in evidence acquisition and answer selection. A checkpoint-wise behavioral diagnostic on CWQ also finds higher late-training recovery rates for OpCredit than for BOC-only after backward operations (Appendix~\ref{app:behavior}). Appendix~\ref{app:error_analysis} provides the protocol, category definitions, and complete statistics.

\section{Conclusion}

We presented OpCredit, an operation-aware policy-learning framework for agentic KGQA. BOC compares terminal task rewards from ordinary rollouts under matched navigation contexts, while FTC measures immediate answer-conditioned frontier changes. Both signals supervise the same decision span during training; inference uses only the learned policy and standard KG interface. Results across three benchmarks and three backbones show improved KGQA performance; controlled ablations support complementary contributions from both signals, and the audit shows closer credit alignment with assessed operation usefulness on a shared \mbox{rollout pool}.

\section*{Limitations}

Our evaluation is limited to three Freebase-based KGQA benchmarks with a fixed executable-operation interface. Performance on other graph schemas and tool environments remains to be evaluated. We focus on operation-aware policy learning within this interface; comparisons of alternative agent interfaces are left for future work.

Our experiments do not cover multilingual or cross-lingual KGQA. Further evaluation could examine questions expressed in different languages, including settings where the question language differs from that of KG entity and relation labels.

We also evaluate against a fixed KG snapshot. The current experiments do not assess how a trained policy responds to subsequent additions or updates to the graph. Evaluation on evolving KGs would complement the present benchmark results.

\bibliography{references}

\clearpage
\appendix
\input{appendix}

\end{document}

%% file: appendix.tex

\section{Experimental Protocol}
\label{app:protocol}

\subsection{Hardware and software}
All experiments were run on a single node equipped with eight NVIDIA RTX 5880 Ada
GPUs (48 GB each). Reinforcement learning (RL) used tensor parallelism of one and
four-way data parallelism; evaluation used single-GPU multi-process workers.
The total compute for SFT, RL training, ablations, and evaluation is estimated
at approximately 2,500 GPU hours.
The software environment used Python 3.10, PyTorch 2.11.0,
Transformers 5.5.4, veRL 0.7.1, and vLLM 0.21.0. The knowledge graph
(KG) was built from the final Freebase RDF dump (2015-08-09) and accessed
through a local SPARQL endpoint.

\subsection{Task, data, and metrics}
\label{app:evaluation}
We evaluate CWQ and WebQSP on their test splits and GrailQA on its full
official development split. Dataset statistics and preprocessing are described in
Appendix~\ref{app:data_stats}. Hit@1 records whether the predicted answer set
contains a gold answer; F1 compares the predicted and gold answer
sets after answer normalization. Formal evaluation uses one sampled
trajectory per question. The audit and behavioral analyses use CWQ
diagnostic subsets with multiple rollouts per question; their sampling settings
are listed separately in Table~\ref{tab:app_hparams}.

\paragraph{Reported baseline scores.}
Table~\ref{tab:main_results} retains the published scores for the named KGQA baselines. For KnowCoder-A1, we report its published F1 values with the
Qwen2.5-Coder-7B backbone. Its CWQ EM and WebQSP Random Hits@1 are not entered
in the Hit@1 columns; its GrailQA overall score of 80.5 is F1. For GAPD, we report its published entity-level F1 on WebQSP and the full GrailQA development split; its exact-set-match EM is not entered in the Hit@1 columns.

\subsection{Training configurations}
We use the instruction-tuned checkpoints \texttt{Qwen/Qwen2.5-3B-Instruct},
\texttt{Qwen/Qwen2.5-7B-Instruct}, and
\texttt{meta-llama/Llama-3.1-8B-Instruct}, each initialized through SFT. The ablation study in Table~\ref{tab:ablation} holds the SFT
initialization, rollout protocol, and optimization settings fixed while
enabling BOC, FTC, both, or neither; the configuration with neither signal
is GRPO. Following the checkpoint-selection protocol of EoG~\citep{eog},
we evaluate the final RL checkpoint unless training diverges, in which case
we use the most recent stable checkpoint according to the training reward curve.
For our configurations, Tables~\ref{tab:main_results} and~\ref{tab:ablation}
report results from selected runs.

\subsection{Hyperparameters}
\label{app:hparams}
Table~\ref{tab:app_hparams} summarizes training, formal evaluation, and
sampling for the audit and behavioral diagnostics. RL training batch sizes count
questions before sampling six rollouts per question: 64 questions yield 384
trajectories for Qwen2.5-3B, and 16 yield 96 for Qwen2.5-7B and Llama-3.1-8B.
The audit and behavioral diagnostic protocols
are specified in Appendices~\ref{app:judge} and~\ref{app:behavior}.

\begin{table*}[t]
\centering
\small
\setlength{\tabcolsep}{5pt}
\renewcommand{\arraystretch}{1.05}
\begin{tabularx}{\textwidth}{@{}l>{\raggedright\arraybackslash}p{0.33\textwidth}>{\raggedright\arraybackslash}X@{}}
\toprule
Stage & Parameter & Setting\\
\midrule
SFT & Effective batch size & 256\\
SFT & Learning rate & $1\times 10^{-5}$\\
SFT & Epochs & 8 (CWQ/WebQSP); 6 (GrailQA)\\
SFT & Maximum sequence length & 5120\\
SFT & Precision and sharding & bf16; DeepSpeed ZeRO-2\\
\midrule
RL & Optimizer learning rate & $1\times 10^{-6}$\\
RL & Training batch size (questions) & 64 (Qwen2.5-3B); 16 (Qwen2.5-7B/Llama-3.1-8B)\\
RL & PPO minibatch size (configured) & 64 (Qwen2.5-3B); 16 (Qwen2.5-7B/Llama-3.1-8B)\\
RL & Rollouts per question & 6\\
RL & Rollout temperature & 1.0\\
RL & Maximum prompt length & 2048\\
RL & Maximum response length & 4096\\
RL & Maximum navigation hops & 10\\
RL & KL coefficient $\beta$ & 0.002\\
RL & Clipping & ratio 0.2 (low/high); dual-clip $c_{\mathrm{dc}}=3.0$\\
RL & Entropy coefficient & 0.001\\
RL & Training steps & 1600; checkpoint every 100 steps\\
RL & Terminal task reward & $0.5\,\mathrm{Hit@1}+0.5\,\mathrm{F1}$\\
\midrule
OpCredit & FTC auxiliary weight $\lambda_{\mathrm{aux}}$ & 0.1\\
OpCredit & FTC eligibility & valid operation-local before/after frontier comparison\\
OpCredit & Numerical stability constant $\varepsilon$ & $10^{-6}$\\
OpCredit & Outcome normalization & group standard deviation for GRPO and BOC; FTC is an independent signed term\\
\midrule
Evaluation & Sampled trajectories per question & 1\\
Evaluation & Sampling temperature & 0.8\\
Evaluation & Maximum navigation hops & 10\\
\midrule
Audit/behavior & Rollouts per question & 8\\
Audit/behavior & Sampling temperature & 1.0\\
Audit/behavior & Maximum navigation hops & 10\\
\bottomrule
\end{tabularx}
\caption{Training and evaluation settings, including rollout sampling for the CWQ credit audit and behavioral diagnostic.}
\label{tab:app_hparams}
\end{table*}

\section{Dataset Statistics}
\label{app:data_stats}

WebQSP provides natural-language questions with Freebase semantic parses; CWQ extends WebQSP with compositional questions; GrailQA covers i.i.d., compositional, and zero-shot generalization. The official releases are available from Microsoft Research for WebQSP,\footnote{\url{https://www.microsoft.com/en-us/download/details.aspx?id=52763}} the authors' project page for CWQ,\footnote{\url{https://www.tau-nlp.sites.tau.ac.il/compwebq}} and the GrailQA project page.\footnote{\url{https://dki-lab.github.io/GrailQA/}} GrailQA is released under CC BY-SA 4.0. Dataset files and pretrained model weights are obtained separately from their respective providers and are not included in the code repository.

Table~\ref{tab:app_data_stats} summarizes the source dataset split sizes. The CWQ/WebQSP SFT--RL pool
excludes empty gold answers or topic entities and training examples sharing
an original source question with any evaluation example. Deduplication leaves
27,237 questions, distinct from the raw split sizes in the table. CWQ records with missing gold answers or topic entities are omitted from scoring. Each example
stores the topic entity and gold answer entities; GrailQA is trained
separately.

\begin{table}[!htbp]
\centering
\small
\setlength{\tabcolsep}{4pt}
\renewcommand{\arraystretch}{1.08}
\begin{tabular*}{\columnwidth}{@{\extracolsep{\fill}}lrr@{}}
\toprule
Dataset & Train & Eval.\\
\midrule
WebQSP & 3,071 & 1,639\\
CWQ & 27,639 & 3,531\\
GrailQA & 44,337 & 6,763\\
\bottomrule
\end{tabular*}
\caption{Source dataset split sizes. Eval. denotes the test split for CWQ/WebQSP and the full development split for GrailQA.}
\label{tab:app_data_stats}
\end{table}

\section{Data Format and Interaction Templates}
\label{app:templates}

\subsection{Agent interaction protocol}
Each decision includes reasoning in a \texttt{<think>} span followed by an
executable operation. Entity-and-direction selection returns the available
relations for the selected entity. Relation selection then returns the
traversed path and reached entity set in an \texttt{<information>} block.
Within a relation-selection operation, unnamed intermediate nodes are expanded
by one additional edge in either direction to return named endpoints;
this expansion counts as one navigation hop in the interaction budget.
The agent continues from these entities or emits its predicted answer set
in an \texttt{<answer>} block.

SFT trajectories are constructed from gold reasoning paths in the benchmark
training data. Templates generate the reasoning text, and each trace is
serialized as reasoning, executable operations, KG observations, and final
answers. All three backbones share the same SFT training data.
Tables~\ref{tab:app_system_prompt} and~\ref{tab:app_user_prompts} give the fixed
system prompt and the user-message templates. Braced fields are filled with
the question or current KG observations. Entity-and-direction selection also
uses the first template at the start of a rollout, with the topic entity as
the current entity. For incoming relation selection, the first line of the
second template is \texttt{Incoming relations (to \{entity\_name\}):}.
Table~\ref{tab:app_interactions} illustrates serialized SFT examples.

\begin{table*}[tp]
\centering
\small
\textbf{System prompt.}\par\smallskip
\begin{tabularx}{\linewidth}{@{}>{\ttfamily\raggedright\arraybackslash}X@{}}
You are a KGQA agent that answers questions by navigating a knowledge graph.
\newline
Explore the graph step by step:
\newline
\mbox{}\newline
1. Choose an entity and a direction to explore:
\newline
\hspace*{3\fontdimen2\font}<explore direction="outgoing">entity\_name</explore> {\rmfamily\textemdash} follow outgoing edges from entity
\newline
\hspace*{3\fontdimen2\font}<explore direction="incoming">entity\_name</explore> {\rmfamily\textemdash} follow incoming edges to entity
\newline
\hspace*{3\fontdimen2\font}Reason in <think > tags before each action.
\newline
\mbox{}\newline
2. After seeing the available relations, select the most relevant one by outputting its name exactly.
\newline
\mbox{}\newline
3. After seeing the resulting entities, either:
\newline
\hspace*{3\fontdimen2\font}- Answer: <answer>["name1", "name2"]</answer> {\rmfamily\textemdash} JSON list of entity names exactly as shown
\newline
\hspace*{3\fontdimen2\font}- Continue: <explore direction="outgoing/incoming">entity\_name</explore>
\newline
\mbox{}\newline
Important:
\newline
- Check ALL entities before answering. Do not pick the first plausible one.
\newline
- If multiple entities satisfy the question, list them all.
\newline
- Do NOT fabricate facts. Only use information visible in names.\\
\end{tabularx}
\caption{Fixed system prompt for the KGQA agent.}
\label{tab:app_system_prompt}
\end{table*}

\begin{table*}[tp]
\centering
\small
\textbf{Entity-and-direction selection.}\par\smallskip
\begin{tabularx}{\linewidth}{@{}>{\ttfamily\raggedright\arraybackslash}X@{}}
Question: \{question\}
\newline
Current entity: \{entity\_name\}
\newline
Explore outgoing or incoming relations from this entity.\\
\end{tabularx}

\medskip
\textbf{Relation selection.}\par\smallskip
\begin{tabularx}{\linewidth}{@{}>{\ttfamily\raggedright\arraybackslash}X@{}}
Outgoing relations (from \{entity\_name\}):
\newline
\hspace*{2\fontdimen2\font}\{relation\_1\}
\newline
\hspace*{2\fontdimen2\font}\{relation\_2\}
\newline
\hspace*{2\fontdimen2\font}...
\newline
\mbox{}\newline
Select the most relevant relation name exactly as shown.\\
\end{tabularx}

\medskip
\textbf{Returned entities.}\par\smallskip
\begin{tabularx}{\linewidth}{@{}>{\ttfamily\raggedright\arraybackslash}X@{}}
<information>
\newline
Question: \{question\}
\newline
Paths explored:
\newline
\hspace*{2\fontdimen2\font}\{head\} \char45{}\char45{}\{relation\}\char45{}\char45{}>
\newline
Entities:
\newline
\hspace*{2\fontdimen2\font}\{entity\_1\}
\newline
\hspace*{2\fontdimen2\font}\{entity\_2\}
\newline
\hspace*{2\fontdimen2\font}...
\newline
</information>
\newline
\mbox{}\newline
Based on the entities above, answer the question or continue exploring. If answering,
\newline
your answer MUST be one of the Entities listed above, copied exactly as shown.\\
\end{tabularx}

\medskip
\textbf{Maximum navigation depth.}\par\smallskip
\begin{tabularx}{\linewidth}{@{}>{\ttfamily\raggedright\arraybackslash}X@{}}
<information>
\newline
Question: \{question\}
\newline
Paths explored:
\newline
\hspace*{2\fontdimen2\font}\{explored\_paths\}
\newline
Entities:
\newline
\hspace*{2\fontdimen2\font}\{entities\}
\newline
</information>
\newline
\mbox{}\newline
You have reached the maximum exploration depth. Provide your final answer in <answer>
\newline
tags as a JSON list of entity names copied exactly from the results above.\\
\end{tabularx}
\caption{User-message templates for KG navigation and final answering.}
\label{tab:app_user_prompts}
\end{table*}

\begin{table*}[tp]
\centering
\small
\setlength{\tabcolsep}{4pt}
\renewcommand{\arraystretch}{1.08}
\textbf{WebQSP SFT trace.}\par\smallskip
\begin{tabularx}{\linewidth}{@{}>{\bfseries\raggedright\arraybackslash}p{0.14\linewidth}>{\raggedright\arraybackslash}X@{}}
Question & what 5 countries border Ethiopia?\\
Assistant & \texttt{<think> I need to inspect outgoing relations for Ethiopia. </think>}\\
 & \texttt{<explore direction="outgoing">Ethiopia</explore>}\\
Observation & \texttt{Outgoing relations: location.location.adjoin\_s, ...}\\
Assistant & \texttt{<think> location.location.adjoin\_s is relevant. </think>}\\
 & \texttt{location.location.adjoin\_s}\\
Observation & \texttt{<information> Ethiopia -> Eritrea, Djibouti, Somalia, Sudan, Kenya </information>}\\
Answer & \texttt{<answer>["Eritrea", "Djibouti", "Somalia", "Sudan", "Kenya"]</answer>}\\
\end{tabularx}

\medskip
\textbf{CWQ SFT trace.}\par\smallskip
\begin{tabularx}{\linewidth}{@{}>{\bfseries\raggedright\arraybackslash}p{0.14\linewidth}>{\raggedright\arraybackslash}X@{}}
Question & Which movies produced by Universal Studios has Ron Howard directed?\\
Assistant & \texttt{<think> I need to inspect outgoing relations for Ron Howard. </think>}\\
 & \texttt{<explore direction="outgoing">Ron Howard</explore>}\\
Observation & \texttt{Outgoing relations: film.director.film, film.producer.film, ...}\\
Assistant & \texttt{<think> film.director.film is relevant. </think>}\\
 & \texttt{film.director.film}\\
Observation & \texttt{<information> Ron Howard -> EDtv, Cinderella Man, A Beautiful Mind, ... </information>}\\
Answer & \texttt{<answer>["EDtv", "Cinderella Man", "A Beautiful Mind"]</answer>}\\
\end{tabularx}
\caption{Serialized SFT examples from WebQSP and CWQ. The examples illustrate the operation and observation format.}
\label{tab:app_interactions}
\end{table*}

\subsection{Recorded rollout information}
Credit computation uses the terminal task reward of each rollout and the
navigation context, executed operation, decision span, and available
before/after frontiers of its decisions. These records align BOC and FTC
with the generated reasoning and executable operation in the same decision
span. Appendix~\ref{app:credit_details} gives the credit calculations and
their alignment with the policy objectives.

\section{Credit Computation and Token Alignment}
\label{app:credit_details}

\subsection{Branch Outcome Credit}
\label{app:boc_example}
For the example in Figure~\ref{fig:overview}, consider three same-question
rollouts $\tau_1$, $\tau_2$, and $\tau_3$ whose operations $o'$, $o_t$, and $o''$ at decision $t$ share the same active entity and
context $c$, with terminal task rewards $R'=0$, $R=1$, and $R''=0$,
respectively. Each operation occurs in one rollout, so Eq.~\eqref{eq:branch_qv} gives
\[
\begin{aligned}
\widehat Q_{\mathrm{out}}(c,o_t)&=1,\\
\widehat Q_{\mathrm{out}}(c,o')=\widehat Q_{\mathrm{out}}(c,o'')&=0,\\
\widehat V_{\mathrm{out}}(c)&=\frac{0+1+0}{3}=\frac13.
\end{aligned}
\]
Substituting these values into Eq.~\eqref{eq:branch_advantage} yields
\[
\begin{aligned}
A_{\mathrm{BOC}}^{2,t}&=\frac{2}{3(\sigma_q+\varepsilon)},\\
A_{\mathrm{BOC}}^{1,t}=A_{\mathrm{BOC}}^{3,t}&=-\frac{1}{3(\sigma_q+\varepsilon)}.
\end{aligned}
\]
Thus $o_t$ receives positive BOC, while $o'$ and $o''$ receive negative BOC.

\subsection{Frontier Transition Credit}
\label{app:ftc_details}

Table~\ref{tab:app_ftc_frontiers} specifies the operation-local frontiers.
An active entity is the entity from which the current navigation hop proceeds;
a singleton frontier contains that entity alone. These comparisons use the
entities involved in each operation, rather than the cumulative set of
retrieved entities.

\begin{table}[!htbp]
\centering
\footnotesize
\setlength{\tabcolsep}{3pt}
\renewcommand{\arraystretch}{1.1}
\begin{tabularx}{\columnwidth}{@{}>{\raggedright\arraybackslash}p{0.29\columnwidth}>{\raggedright\arraybackslash}X>{\raggedright\arraybackslash}X@{}}
\toprule
Operation & Before frontier $F^-$ & After frontier $F^+$\\
\midrule
Entity-and-direction selection & \{previous active entity\} & \{selected entity\}\\
Relation selection & \{active entity\} & Returned entity set\\
Answer emission & \{active entity\} & Predicted answer entity set\\
\bottomrule
\end{tabularx}
\caption{Operation-local frontiers for FTC. Braces denote a singleton entity set.}
\label{tab:app_ftc_frontiers}
\end{table}

To construct the one-hop neighborhood used by $\delta_q$ in
Eq.~\eqref{eq:frontier_potential}, we retrieve bounded incoming and outgoing
KG neighbors of each mapped gold answer entity, exclude deprecated or internal
predicates, and take the union across gold answer entities.
The bounded lookup uses shallow bidirectional neighborhood expansion and set
intersection.
If no gold answer can be mapped to a KG entity, the question receives no
FTC; an entity unresolved within the agent's visible frontier suppresses only that decision's
FTC. At answer emission, the after frontier contains the predicted answer entities and is evaluated using the same maximum-potential aggregation. Empty or unavailable frontiers also receive no FTC.
For nonempty, available frontiers, Eqs.~\eqref{eq:frontier_potential}
and~\eqref{eq:structural_advantage} give the equivalent computation
\begin{equation}
 A_{\mathrm{FTC}}^{i,t}
 =\min_{e\in F_{i,t}^{-}}\delta_q(e)
  -\min_{e\in F_{i,t}^{+}}\delta_q(e).
 \label{eq:app_ftc}
\end{equation}

\subsection{Decision spans and joint objective}
\label{app:joint_details}
We use the decision spans and credit assignments defined in
Section~\ref{sec:policy_learning}.
Each loss term is averaged over generated tokens within a trajectory and then
over trajectories, giving each trajectory equal weight. Observation and padding
tokens are excluded. KL regularization uses the frozen SFT-initialized policy
as its reference; the entropy bonus uses the current policy's token distribution.

\paragraph{BOC surrogate.}
For the policy ratio in Eq.~\eqref{eq:policy_ratio}, define
\[
\begin{aligned}
\bar\rho&=\operatorname{clip}(\rho,1-\eta,1+\eta),\\
m(\rho,A)&=\min\{\rho A,\bar\rho A\}.
\end{aligned}
\]
The dual-clipped surrogate $D$ in $\mathcal L_{\mathrm{BOC}}(\theta)$ is
\begin{equation}
 \begin{aligned}
 D(\rho,A)
 &=m(\rho,A), && A\geq 0,\\
 D(\rho,A)
 &=\max\{m(\rho,A),c_{\mathrm{dc}}A\}, && A<0.
 \end{aligned}
 \label{eq:app_dclip}
\end{equation}
Here $\eta$ is the policy-ratio clipping parameter and $c_{\mathrm{dc}}$
is the dual-clip bound (set to $3.0$ in our runs).

\subsection{Analysis of Navigation Context Matching}
\label{app:context_key}

We examine the role of decision phase in BOC navigation context matching
using CWQ diagnostic rollouts from Qwen2.5-3B with FTC at steps 100, 300,
500, and 600, and Qwen2.5-7B with BOC at steps 100, 300, and 600. Each
checkpoint contributes eight rollouts per question on 500 questions.
The analysis considers relation selection and entity-and-direction selection.
Every matching rule restricts both outcome estimates to the same question and
active entity, as in Section~\ref{sec:branch_credit}. We hold terminal task
rewards fixed and perform no parameter updates.
For each matching rule, we recompute the
context-relative outcome credit using the same normalization as in training.
Within each question--active-entity group, we compare three alternatives
with the full context: (i) active entity only, with no further matching;
(ii) visited entity set only, matching $E_{\mathrm{vis}}(s)$;
and (iii) visited entity set with remaining navigation budget, matching
$(E_{\mathrm{vis}}(s),b(s))$. The full context in
Eq.~\eqref{eq:context_key} additionally matches decision phase $\phi(s)$.
The three alternatives without decision phase produce identical partitions
on these rollouts and are grouped in one row in
Table~\ref{tab:context_key_mixing}.

The decision-weighted mixing rate is the fraction of decision occurrences
in context groups containing both relation selection and
entity-and-direction selection. The group-weighted rate is the fraction of
context groups containing both operation types.

\begin{table}[t]
\centering
\footnotesize
\setlength{\tabcolsep}{4pt}
\renewcommand{\arraystretch}{1.08}
\begin{tabular*}{\columnwidth}{@{\extracolsep{\fill}}lcc@{}}
\toprule
\textbf{Context} & \shortstack{\textbf{Decision-}\\\textbf{weighted}} & \shortstack{\textbf{Group-}\\\textbf{weighted}} \\
\midrule
Without decision phase & 46--48\% & 30--31\% \\
\rowcolor{blue!10}Full & 0\% & 0\% \\
\bottomrule
\end{tabular*}
\caption{Decision-phase mixing under different navigation context matching rules. Rates are consistent across the evaluated checkpoints.}
\label{tab:context_key_mixing}
\end{table}

The decision-weighted rate measures how many decisions share a context-relative
outcome baseline with a different operation type. Without decision phase, this affects
nearly half of the sampled decision occurrences. Matching the decision phase provides
a baseline formed from the same operation type, aligning the local outcome
comparison with the kind of decision being made. The visited entity set and
remaining navigation budget retain the other matching constraints.

\section{Reference-Blinded Credit Audit}
\label{app:judge}

\subsection{Audit setup}
The audit uses a common rollout pool generated by the Qwen2.5-3B SFT policy
on 500 CWQ questions, with eight rollouts per question. It contains 11,298
operation occurrences; 1,200 are selected for usefulness assessment after
removing duplicate operation signatures within each question and capping
the number assessed per question.

The compared methods are GRPO, GiGPO, RTMC, GraphGPO, and OpCredit. We
reimplement the published credit-assignment rules for KGQA and compute
their credits on the complete rollout pool before selecting the judged
occurrences. The rollout pool, terminal task rewards, and assessed
operations are held fixed across methods.

\subsection{Reference-blinded usefulness assessment}
We use Qwen3.7-Max as the reference-blinded judge with temperature 0.
Given the question, interaction history preceding the operation, the
operation itself, and the graph observations it returns, the judge provides
a usefulness rating and a one-sentence reason. It does not receive the
gold answer, terminal task reward, answer-distance label, assigned credit,
or later trajectory steps.
Table~\ref{tab:app_judge_prompt} gives the prompt and five-point annotation
rubric. For mean-credit visualization, ratings 2 and 3 share one group,
yielding the four ordered levels in Figure~\ref{fig:credit_alignment}.
Table~\ref{tab:app_audit_labels} relates the original ratings to the displayed
levels and AUC labels.

\begin{table}[!htbp]
\centering
\footnotesize
\setlength{\tabcolsep}{5pt}
\renewcommand{\arraystretch}{1.05}
\begin{tabular*}{\columnwidth}{@{\extracolsep{\fill}}ccc@{}}
\toprule
Judge rating & Displayed level & AUC label\\
\midrule
1 & 1 & Unhelpful\\
2 & 2 & Unhelpful\\
3 & 2 & Excluded (neutral)\\
4 & 3 & Useful\\
5 & 4 & Useful\\
\bottomrule
\end{tabular*}
\caption{Annotation ratings, displayed usefulness levels, and AUC labels.}
\label{tab:app_audit_labels}
\end{table}

Truncated relation names were resolved against the KG, and operations
whose visible judge prompts changed were reassessed.

\begin{table*}[t]
\centering
\small
\setlength{\tabcolsep}{6pt}
\renewcommand{\arraystretch}{1.15}
\begin{tabular}{@{}p{0.15\textwidth}p{0.80\textwidth}@{}}
\toprule
\textbf{Message} & \textbf{Content}\\
\midrule
System &
\begin{minipage}[t]{0.80\textwidth}\raggedright
You are an expert evaluator for knowledge-graph navigation steps.  Given a
question and an agent's navigation trajectory on a Freebase-style knowledge
graph, rate one operation for how much it advances answering the question.
You do not know the correct answer: judge purely from the semantics of the
question, the operation description, and the entities involved.  Use the following
scale: 5 = clearly moves toward the answer; 4 = probably helpful; 3 = plausible
exploration, neutral, or uncertain; 2 = probably unhelpful; 1 = clearly
off-topic or counterproductive.  Respond with only a JSON object of the form
\texttt{\{"rating": <1--5>, "reason": "<one sentence>"\}}.
\end{minipage}\\
User &
\begin{minipage}[t]{0.80\textwidth}\raggedright
Question: \{question\}.  Navigation path so far:
\{path-before-the-operation\}.  Operation to evaluate:
\{operation description\}.  Rate this single operation from 1 to 5.
\end{minipage}\\
\bottomrule
\end{tabular}
\par\smallskip
\parbox{\textwidth}{\footnotesize\textbf{Graph observations.} The observations returned by the evaluated operation are included in \texttt{\{operation description\}}.}
\caption{Reference-blinded judge prompt.}
\label{tab:app_judge_prompt}
\end{table*}

\subsection{Credit metrics and results}
For OpCredit, the replay credit on an eligible operation is
\begin{equation}
 A_{\mathrm{replay}}=A_{\mathrm{BOC}}+\lambda_{\mathrm{aux}}A_{\mathrm{FTC}},
 \label{eq:app_combined}
\end{equation}
where $\lambda_{\mathrm{aux}}=0.1$ and $A_{\mathrm{BOC}}$ follows the assignment
in Eq.~\eqref{eq:branch_advantage}, with rollout and decision indices omitted.
With policy ratio one and fixed aggregation, this is the
operation-level policy-gradient coefficient excluding KL and entropy terms.

For each method, we record the mean replay credit in each usefulness level and the usefulness-discrimination AUC. AUC uses assigned credit as the score, with original ratings 4--5 as positive labels and 1--2 as negative labels; rating 3 is excluded. All methods use the same labeled operations. AUC is computed from individual-operation credit rankings before grouping ratings for the mean-credit display. Mean credit describes the direction of supervision within each method, while AUC compares usefulness discrimination without relying on a common credit scale. Table~\ref{tab:app_credit_comparison} reports the complete results.

\begin{table}[!htbp]
\centering
\footnotesize
\setlength{\tabcolsep}{3pt}
\renewcommand{\arraystretch}{1.1}
\begin{tabular}{lrrrrr}
\toprule
Method & Level 1 & Level 2 & Level 3 & Level 4 & AUC\\
\midrule
GRPO & $-0.134$ & $-0.019$ & $+0.005$ & $+0.014$ & 0.606\\
GiGPO & $-0.556$ & $-0.003$ & $-0.056$ & $+0.087$ & 0.592\\
RTMC & $-0.108$ & $-0.031$ & $-0.016$ & $+0.003$ & 0.492\\
GraphGPO & $-0.471$ & $+0.000$ & $-0.020$ & $+0.057$ & 0.595\\
\opcredit & $-0.347$ & $+0.016$ & $+0.050$ & $+0.144$ & \textbf{0.718}\\
\bottomrule
\end{tabular}
\caption{Credit-assignment replay on the common SFT rollout pool. Mean credit is reported for the four ordered usefulness levels; AUC measures usefulness discrimination.}
\label{tab:app_credit_comparison}
\end{table}

On the shared rollout pool, OpCredit, GRPO, and RTMC all show increasing mean credit across usefulness levels. OpCredit achieves the highest AUC, indicating stronger discriminability between useful and unhelpful operations.

\section{Behavioral Diagnostic}
\label{app:behavior}

\begin{figure*}[t]
\centering
\includegraphics[width=\textwidth]{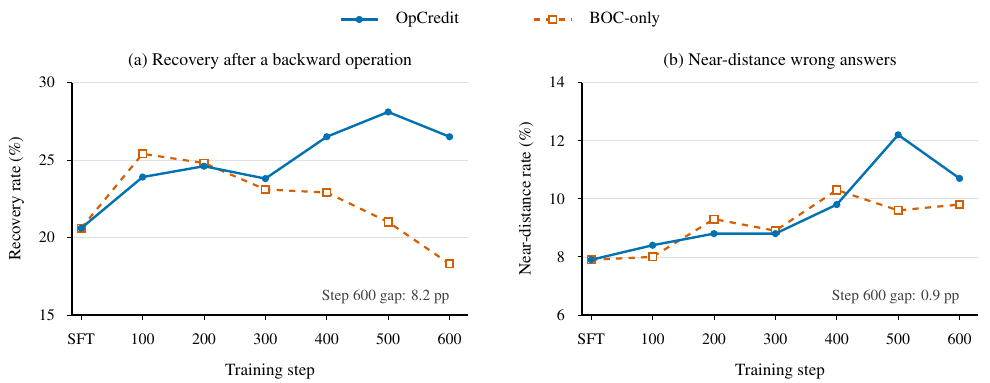}
\caption{Behavioral diagnostic on CWQ for Qwen2.5-3B. The recovery rate is
conditioned on trajectories containing a backward operation ($\Delta d_{\mathrm{diag}}>0$);
the near-distance rate is conditioned on wrong final answers. Values are
single-run percentages over the fixed diagnostic protocol. OpCredit rises to
the highest late-training recovery rate, while the two methods have smaller,
less consistent differences in near-distance errors.}
\label{fig:behavior_trends}
\end{figure*}

\begin{table}[!htbp]
\centering
\footnotesize
\setlength{\tabcolsep}{1.2pt}
\renewcommand{\arraystretch}{1.08}
\begin{tabular}{@{}lrrrrrrr@{}}
\toprule
Metric & SFT & 100 & 200 & 300 & 400 & 500 & 600\\
\midrule
Near-distance, OpCredit & 7.9 & 8.4 & 8.8 & 8.8 & 9.8 & 12.2 & 10.7\\
Near-distance, BOC-only & 7.9 & 8.0 & 9.3 & 8.9 & 10.3 & 9.6 & 9.8\\
Recovery, OpCredit & 20.6 & 23.9 & 24.6 & 23.8 & 26.5 & \textbf{28.1} & 26.5\\
Recovery, BOC-only & 20.6 & \textbf{25.4} & 24.8 & 23.1 & 22.9 & 21.0 & 18.3\\
\bottomrule
\end{tabular}
\caption{Behavioral diagnostic on CWQ. Values are percentages; BOC-only denotes OpCredit without FTC.}
\label{tab:app_behavior}
\end{table}

\subsection{Diagnostic setup and metrics}
We compare Qwen2.5-3B OpCredit with BOC-only (OpCredit without
FTC) on 500 CWQ questions sampled with seed 42, using the diagnostic rollout
settings in Table~\ref{tab:app_hparams}. Each configuration is evaluated at
SFT and six training checkpoints (100, 200, 300, 400, 500, and 600).

For each operation, the diagnostic records a four-level diagnostic distance
label $d_{\mathrm{diag}}$: 0 for a gold answer entity, 1 for the constructed one-hop neighborhood
of the gold answer entities, 2 for the restricted second-level neighborhood, and 3 for a
more distant candidate.  The diagnostic distinguishes second-level and more distant candidates
within FTC label 2 in Eq.~\eqref{eq:frontier_potential}.

The \emph{wrong-answer near-distance rate} is
\[
 P\bigl(d_{\mathrm{answer}}\leq 1\mid
        \text{final answer is wrong}\bigr).
\]
Here $d_{\mathrm{answer}}$ is the diagnostic distance label of the predicted answer. The denominator is the set of wrong-answer trajectories, and the metric asks
whether an incorrect answer is nevertheless in the answer neighborhood.  The
\emph{recovery rate} is a trajectory-level conditional probability:
\[
 P\bigl(\text{correct}\mid\text{trajectory has backward op.}\bigr),
\]
where a backward operation is a navigation operation with
$\Delta d_{\mathrm{diag}}=d_{\mathrm{diag}}^{+}-d_{\mathrm{diag}}^{-}>0$.
The denominator contains trajectories with at least one such operation.

\paragraph{Illustrative recovery.}
For ``What all has the actor who played in Knights of the Roundtable:
King Arthur been in?'', the recorded before/after $d_{\mathrm{diag}}$ pairs
for four operations are
$0\!\rightarrow\!3$, $0\!\rightarrow\!3$, $3\!\rightarrow\!0$, and
$3\!\rightarrow\!0$. Each pair describes an operation-local comparison, so adjacent pairs need not share an endpoint. The trajectory first moves away from the answer region,
then changes relation and returns to it before producing the correct answer.

\subsection{Seven-checkpoint results}
Figure~\ref{fig:behavior_trends} and Table~\ref{tab:app_behavior} report the
checkpoint-wise results. OpCredit maintains higher recovery rates than
BOC-only at the later checkpoints, reaching 26.5\% versus 18.3\%
at step 600. This comparison concerns trajectories that contain a backward
operation, locating the difference in recovery following such operations.
The wrong-answer near-distance rate shows smaller differences with no
consistent direction across checkpoints. The clearest distinction between
the configurations is therefore in recovery after backward operations.

\section{Error Analysis Details}
\label{app:error_analysis}

\subsection{Diagnostic protocol}
We compare GRPO and OpCredit on 492 matched diagnostic questions per
dataset using the agent's executable KG interface. Errors are summarized
at the question level on these diagnostic subsets. Figure~\ref{fig:error_analysis}
reports category shares within each method's error set.

\subsection{Error categories}
The automated diagnostic assigns four labels based on candidate availability, reference-path alignment, and answer emission:
\begin{itemize}
    \item \textbf{Answer selection.} An incorrect answer receives this label when a gold-answer name occurs in the final frontier or the complete path aligns with a reference path. Complete-path alignment serves as a proxy for reaching relevant evidence.
    \item \textbf{Path-first.} The recorded deviation from a reference path occurs at the first relation hop.
    \item \textbf{Path-late.} The recorded deviation occurs at the second or a later hop, or the unsuccessful trajectory has no usable reference-path alignment. This category therefore includes unaligned residual cases as well as localized later deviations.
    \item \textbf{Empty answer.} The predicted answer set is empty.
\end{itemize}
Path deviations are defined relative to the available reference paths.

\subsection{Complete diagnostic statistics}
Table~\ref{tab:app_error_analysis} reports error counts and rounded category shares. The percentages may not sum to 100\%; Figure~\ref{fig:error_analysis} preserves these labels and normalizes the bar widths.

\begin{table}[!htbp]
\centering
\small
\setlength{\tabcolsep}{3.1pt}
\renewcommand{\arraystretch}{1.12}
\begin{tabular}{@{}lrrrrr@{}}
\toprule
\textbf{Method} & \textbf{Err.} & \textbf{Sel.} & \textbf{First} & \textbf{Late} & \textbf{Empty} \\
\midrule
\multicolumn{6}{@{}l}{\textit{CWQ}} \\
GRPO & 105 & 67 & 23 & 10 & 0 \\
OpCredit & 97 & 61 & 23 & 16 & 0 \\
\midrule
\multicolumn{6}{@{}l}{\textit{WebQSP}} \\
GRPO & 55 & 27 & 22 & 35 & 16 \\
OpCredit & 49 & 20 & 24 & 33 & 22 \\
\midrule
\multicolumn{6}{@{}l}{\textit{GrailQA}} \\
GRPO & 107 & 31 & 47 & 14 & 8 \\
OpCredit & 95 & 23 & 51 & 20 & 6 \\
\bottomrule
\end{tabular}
\caption{Total error counts (Err.) and within-method category percentages on 492 diagnostic questions per dataset. Sel.: Answer selection; First: Path-first; Late: Path-late; Empty: Empty answer.}
\label{tab:app_error_analysis}
\end{table}

%% file: main.bbl
\begin{thebibliography}{25}
\providecommand{\natexlab}[1]{#1}

\bibitem[{Chen et~al.(2026{\natexlab{a}})Chen, Liu, Xie, and Xiong}]{scprm}
Jiujiu Chen, Yazheng Liu, Sihong Xie, and Hui Xiong. 2026{\natexlab{a}}.
\newblock \href {https://doi.org/10.48550/ARXIV.2605.02819} {{SCPRM:} {A} schema-aware cumulative process reward model for knowledge graph question answering}.
\newblock \emph{CoRR}, abs/2605.02819.

\bibitem[{Chen et~al.(2024)Chen, Tong, Jin, Sun, Ye, and Xiong}]{pog}
Liyi Chen, Panrong Tong, Zhongming Jin, Ying Sun, Jieping Ye, and Hui Xiong. 2024.
\newblock \href {http://papers.nips.cc/paper\_files/paper/2024/hash/4254e856d01a5e7b7ea050477c3ef9b9-Abstract-Conference.html} {Plan-on-graph: Self-correcting adaptive planning of large language model on knowledge graphs}.
\newblock In \emph{Advances in Neural Information Processing Systems 37: Annual Conference on Neural Information Processing Systems 2024, NeurIPS 2024, Vancouver, BC, Canada, December 10 - 15, 2024}.

\bibitem[{Chen et~al.(2026{\natexlab{b}})Chen, Wang, Li, Zhang, Ding, Yang, Xu, and Jin}]{knowcodera1}
Zhuo Chen, Fei Wang, Zixuan Li, Zhao Zhang, Weiwei Ding, Chuanguang Yang, Yongjun Xu, and Xiaolong Jin. 2026{\natexlab{b}}.
\newblock \href {https://doi.org/10.1145/3774904.3792662} {Incentivizing agentic reasoning capability with outcome supervision for knowledge base question answering}.
\newblock In \emph{Proceedings of the {ACM} Web Conference 2026, {WWW} 2026, Dubai, United Arab Emirates, originally scheduled for April 13-17, 2026, rescheduled for June 29 - July 3, 2026}, pages 4302--4313. {ACM}.

\bibitem[{Cheng et~al.(2026)Cheng, He, Feng, Xu, Yan, Feng, and An}]{graphgpo}
Xin Cheng, Shuo He, Lang Feng, Haiyang Xu, Ming Yan, Lei Feng, and Bo~An. 2026.
\newblock \href {https://doi.org/10.48550/ARXIV.2605.26684} {Beyond trajectory-level attribution: Graph-based credit assignment for agentic reinforcement learning}.
\newblock \emph{CoRR}, abs/2605.26684.

\bibitem[{Feng et~al.(2025)Feng, Xue, Liu, and An}]{gigpo}
Lang Feng, Zhenghai Xue, Tingcong Liu, and Bo~An. 2025.
\newblock \href {http://papers.nips.cc/paper\_files/paper/2025/hash/420c9f777c0b4f78d515e53cf74d58b2-Abstract-Conference.html} {Group-in-group policy optimization for {LLM} agent training}.
\newblock In \emph{Advances in Neural Information Processing Systems 38: Annual Conference on Neural Information Processing Systems 2025, NeurIPS 2025, San Diego, CA, USA, December 2-7, 2025 / Mexico City, Mexico, November 30 - December 5, 2025}.

\bibitem[{Gu et~al.(2021)Gu, Kase, Vanni, Sadler, Liang, Yan, and Su}]{grailqa}
Yu~Gu, Sue Kase, Michelle Vanni, Brian~M. Sadler, Percy Liang, Xifeng Yan, and Yu~Su. 2021.
\newblock \href {https://doi.org/10.1145/3442381.3449992} {Beyond {I.I.D.:} three levels of generalization for question answering on knowledge bases}.
\newblock In \emph{{WWW} '21: The Web Conference 2021, Virtual Event / Ljubljana, Slovenia, April 19-23, 2021}, pages 3477--3488. {ACM} / {IW3C2}.

\bibitem[{Jiang et~al.(2025)Jiang, Zhou, Zhao, Song, Zhu, Zhu, and Wen}]{kgagent}
Jinhao Jiang, Kun Zhou, Xin Zhao, Yang Song, Chen Zhu, Hengshu Zhu, and Ji{-}Rong Wen. 2025.
\newblock \href {https://doi.org/10.18653/V1/2025.ACL-LONG.468} {Kg-agent: An efficient autonomous agent framework for complex reasoning over knowledge graph}.
\newblock In \emph{Proceedings of the 63rd Annual Meeting of the Association for Computational Linguistics (Volume 1: Long Papers), {ACL} 2025, Vienna, Austria, July 27 - August 1, 2025}, pages 9505--9523. Association for Computational Linguistics.

\bibitem[{Li et~al.(2025)Li, Zhang, Wu, Luo, Glass, and Meng}]{dog}
Kun Li, Tianhua Zhang, Xixin Wu, Hongyin Luo, James~R. Glass, and Helen~M. Meng. 2025.
\newblock \href {https://doi.org/10.18653/V1/2025.ACL-LONG.1186} {Decoding on graphs: Faithful and sound reasoning on knowledge graphs through generation of well-formed chains}.
\newblock In \emph{Proceedings of the 63rd Annual Meeting of the Association for Computational Linguistics (Volume 1: Long Papers), {ACL} 2025, Vienna, Austria, July 27 - August 1, 2025}, pages 24349--24364. Association for Computational Linguistics.

\bibitem[{Lin et~al.(2025)Lin, Liu, Song, Wang, Shun, and Zhu}]{kgr1}
Junhong Lin, Shicheng Liu, Jinyeop Song, Song Wang, Julian Shun, and Yada Zhu. 2025.
\newblock \href {https://doi.org/10.48550/ARXIV.2509.26383} {Efficient and transferable agentic knowledge graph {RAG} via reinforcement learning}.
\newblock \emph{CoRR}, abs/2509.26383.

\bibitem[{Luo et~al.(2025)Luo, E, Guo, Lin, Wu, Mu, Liu, Song, Zhu, and Luu}]{kbqao1}
Haoran Luo, Haihong E, Yikai Guo, Qika Lin, Xiaobao Wu, Xinyu Mu, Wenhao Liu, Meina Song, Yifan Zhu, and Anh~Tuan Luu. 2025.
\newblock \href {https://proceedings.mlr.press/v267/luo25d.html} {Kbqa-o1: Agentic knowledge base question answering with monte carlo tree search}.
\newblock In \emph{Forty-second International Conference on Machine Learning, {ICML} 2025, Vancouver, BC, Canada, July 13-19, 2025}, volume 267 of \emph{Proceedings of Machine Learning Research}. {PMLR} / OpenReview.net.

\bibitem[{Luo et~al.(2024)Luo, Li, Haffari, and Pan}]{rog}
Linhao Luo, Yuan{-}Fang Li, Gholamreza Haffari, and Shirui Pan. 2024.
\newblock \href {https://openreview.net/forum?id=ZGNWW7xZ6Q} {Reasoning on graphs: Faithful and interpretable large language model reasoning}.
\newblock In \emph{The Twelfth International Conference on Learning Representations, {ICLR} 2024, Vienna, Austria, May 7-11, 2024}. OpenReview.net.

\bibitem[{Mavromatis and Karypis(2025)}]{gnnrag}
Costas Mavromatis and George Karypis. 2025.
\newblock \href {https://doi.org/10.18653/V1/2025.FINDINGS-ACL.856} {{GNN-RAG:} graph neural retrieval for efficient large language model reasoning on knowledge graphs}.
\newblock In \emph{Findings of the Association for Computational Linguistics, {ACL} 2025, Vienna, Austria, July 27 - August 1, 2025}, volume {ACL} 2025 of \emph{Findings of {ACL}}, pages 16682--16699. Association for Computational Linguistics.

\bibitem[{Ng et~al.(1999)Ng, Harada, and Russell}]{ng1999policy}
Andrew~Y. Ng, Daishi Harada, and Stuart Russell. 1999.
\newblock Policy invariance under reward transformations: Theory and application to reward shaping.
\newblock In \emph{Proceedings of the Sixteenth International Conference on Machine Learning {(ICML} 1999), Bled, Slovenia, June 27 - 30, 1999}, pages 278--287. Morgan Kaufmann.

\bibitem[{Shao et~al.(2024)Shao, Wang, Zhu, Xu, Song, Zhang, Li, Wu, and Guo}]{grpo}
Zhihong Shao, Peiyi Wang, Qihao Zhu, Runxin Xu, Junxiao Song, Mingchuan Zhang, Y.~K. Li, Y.~Wu, and Daya Guo. 2024.
\newblock \href {https://doi.org/10.48550/ARXIV.2402.03300} {Deepseekmath: Pushing the limits of mathematical reasoning in open language models}.
\newblock \emph{CoRR}, abs/2402.03300.

\bibitem[{Song et~al.(2026{\natexlab{a}})Song, Liu, Guti{\'{e}}rrez{-}Basulto, Hu, Xie, Peng, Ananiadou, and Pan}]{graphrft}
Yanlin Song, Ben Liu, V{\'{\i}}ctor Guti{\'{e}}rrez{-}Basulto, Zhiwei Hu, Qianqian Xie, Min Peng, Sophia Ananiadou, and Jeff~Z. Pan. 2026{\natexlab{a}}.
\newblock \href {https://doi.org/10.1145/3774904.3792191} {Plan then retrieve: Reinforcement learning-guided complex reasoning over knowledge graphs}.
\newblock In \emph{Proceedings of the {ACM} Web Conference 2026, {WWW} 2026, Dubai, United Arab Emirates, originally scheduled for April 13-17, 2026, rescheduled for June 29 - July 3, 2026}, pages 3666--3676. {ACM}.

\bibitem[{Song et~al.(2026{\natexlab{b}})Song, Shen, Yu, and Xia}]{dtpo}
Yurun Song, Xiangqing Shen, Jianfei Yu, and Rui Xia. 2026{\natexlab{b}}.
\newblock \href {https://doi.org/10.18653/V1/2026.FINDINGS-ACL.1023} {Interactive semantic parsing with reinforcement learning for knowledge graph reasoning}.
\newblock In \emph{Findings of the Association for Computational Linguistics, {ACL} 2026, San Diego, California, United States, July 2-7, 2026}, pages 20464--20484. Association for Computational Linguistics.

\bibitem[{Sun et~al.(2024)Sun, Xu, Tang, Wang, Lin, Gong, Ni, Shum, and Guo}]{tog}
Jiashuo Sun, Chengjin Xu, Lumingyuan Tang, Saizhuo Wang, Chen Lin, Yeyun Gong, Lionel~M. Ni, Heung{-}Yeung Shum, and Jian Guo. 2024.
\newblock \href {https://openreview.net/forum?id=nnVO1PvbTv} {Think-on-graph: Deep and responsible reasoning of large language model on knowledge graph}.
\newblock In \emph{The Twelfth International Conference on Learning Representations, {ICLR} 2024, Vienna, Austria, May 7-11, 2024}. OpenReview.net.

\bibitem[{Sun et~al.(2026)Sun, Xie, Chen, Liu, Wu, Song, Wang, Wang, and Wang}]{gapd}
Xin Sun, Jianan Xie, Zhongqi Chen, Qiang Liu, Shu Wu, Bowen Song, Weiqiang Wang, Zilei Wang, and Liang Wang. 2026.
\newblock \href {https://doi.org/10.48550/ARXIV.2605.29584} {{GAPD:} gold-action policy distillation for agentic reinforcement learning in knowledge base question answering}.
\newblock \emph{CoRR}, abs/2605.29584.

\bibitem[{Talmor and Berant(2018)}]{cwq}
Alon Talmor and Jonathan Berant. 2018.
\newblock \href {https://doi.org/10.18653/V1/N18-1059} {The web as a knowledge-base for answering complex questions}.
\newblock In \emph{Proceedings of the 2018 Conference of the North American Chapter of the Association for Computational Linguistics: Human Language Technologies, {NAACL-HLT} 2018, New Orleans, Louisiana, USA, June 1-6, 2018, Volume 1 (Long Papers)}, pages 641--651. Association for Computational Linguistics.

\bibitem[{Wang et~al.(2025)Wang, Lin, Guo, Shun, Li, and Zhu}]{reknos}
Song Wang, Junhong Lin, Xiaojie Guo, Julian Shun, Jundong Li, and Yada Zhu. 2025.
\newblock \href {https://openreview.net/forum?id=rTCJ29pkuA} {Reasoning of large language models over knowledge graphs with super-relations}.
\newblock In \emph{The Thirteenth International Conference on Learning Representations, {ICLR} 2025, Singapore, April 24-28, 2025}. OpenReview.net.

\bibitem[{Wang et~al.(2026{\natexlab{a}})Wang, Zheng, and Xu}]{rtmc}
Tao Wang, Suhang Zheng, and Xiaoxiao Xu. 2026{\natexlab{a}}.
\newblock \href {https://doi.org/10.48550/ARXIV.2604.11037} {{RTMC:} step-level credit assignment via rollout trees}.
\newblock \emph{CoRR}, abs/2604.11037.

\bibitem[{Wang et~al.(2026{\natexlab{b}})Wang, Song, Tian, Liu, Luo, and Huang}]{dprm}
Xinyi Wang, Yiping Song, Zhiliang Tian, Bo~Liu, Tingjin Luo, and Minlie Huang. 2026{\natexlab{b}}.
\newblock \href {https://doi.org/10.1609/AAAI.V40I40.40658} {{DPRM:} {A} dual implicit process reward model in multi-hop question answering}.
\newblock In \emph{Fortieth {AAAI} Conference on Artificial Intelligence, Thirty-Eighth Conference on Innovative Applications of Artificial Intelligence, Sixteenth Symposium on Educational Advances in Artificial Intelligence, {AAAI} 2026, Singapore, January 20-27, 2026}, pages 33683--33691. {AAAI} Press.

\bibitem[{Xu et~al.(2026)Xu, Xu, Liu, Yuan, Peng, Zhao, and Liu}]{graphwalker}
Shuwen Xu, Yao Xu, Jiaxiang Liu, Chenhao Yuan, Wenshuo Peng, Jun Zhao, and Kang Liu. 2026.
\newblock \href {https://doi.org/10.48550/ARXIV.2603.28533} {Graphwalker: Agentic knowledge graph question answering via synthetic trajectory curriculum}.
\newblock \emph{CoRR}, abs/2603.28533.

\bibitem[{Yan et~al.(2026)Yan, Chen, Zhou, Yao, Chen, Zhang, Zhang, Zhang, Huang, Duan, and Zhang}]{eog}
Shiqi Yan, Yubo Chen, Ruiqi Zhou, Zhengxi Yao, Shuai Chen, Tianyi Zhang, Shijie Zhang, Wei~Qiang Zhang, Yongfeng Huang, Haixin Duan, and Yunqi Zhang. 2026.
\newblock \href {https://proceedings.iclr.cc/paper\_files/paper/2026/hash/5d2e24df9cfaad3189833b819c40b392-Abstract-Conference.html} {Explore-on-graph: Incentivizing autonomous exploration of large language models on knowledge graphs with path-refined reward modeling}.
\newblock In \emph{The Fourteenth International Conference on Learning Representations, {ICLR} 2026, Rio de Janeiro, Brazil, April 23-27, 2026}. proceedings.iclr.cc.

\bibitem[{Yih et~al.(2016)Yih, Richardson, Meek, Chang, and Suh}]{webqsp}
Wen{-}tau Yih, Matthew Richardson, Christopher Meek, Ming{-}Wei Chang, and Jina Suh. 2016.
\newblock \href {https://doi.org/10.18653/V1/P16-2033} {The value of semantic parse labeling for knowledge base question answering}.
\newblock In \emph{Proceedings of the 54th Annual Meeting of the Association for Computational Linguistics, {ACL} 2016, August 7-12, 2016, Berlin, Germany, Volume 2: Short Papers}. The Association for Computer Linguistics.

\end{thebibliography}
